\documentclass[10pt,twocolumn,letterpaper]{article}

\usepackage{cvpr}
\usepackage{times}
\usepackage{epsfig}
\usepackage{graphicx}
\usepackage{amsmath}
\usepackage{amssymb}
\usepackage{caption}
\usepackage{subcaption}
\usepackage{bm}
\usepackage{multirow}
\usepackage{array}

\usepackage[pagebackref=true,breaklinks=true,letterpaper=true,colorlinks,bookmarks=false]{hyperref}

\cvprfinalcopy 

\def\cvprPaperID{4665} 

\ifcvprfinal\fi
\begin{document}

\title{PBNS: Physically Based Neural Simulator for Unsupervised Garment Pose Space Deformation}

\author{Hugo Bertiche, Meysam Madadi and Sergio Escalera\\
Universitat de Barcelona and Computer Vision Center, Spain\\
{\tt\small hugo\_bertiche@hotmail.com}
}

\twocolumn[{%
\renewcommand\twocolumn[1][]{#1}%
\maketitle
\begin{center}
    \includegraphics[width=\textwidth]{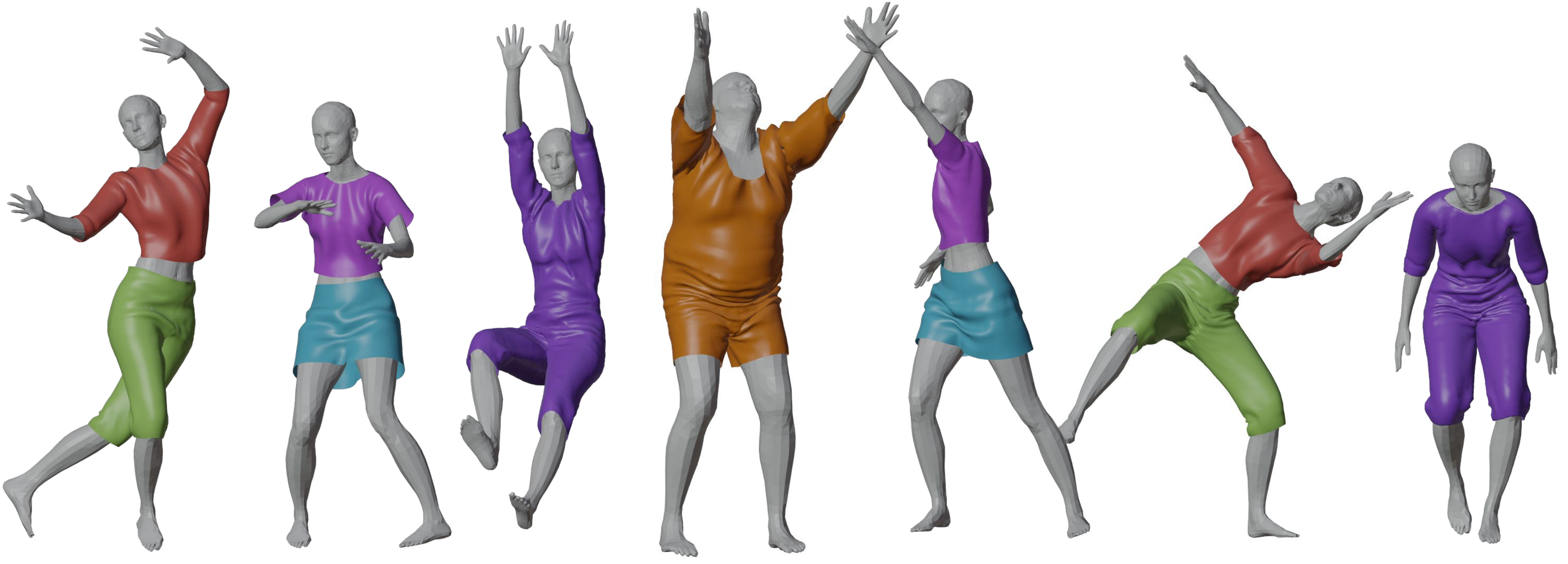}
    \captionof{figure}{We present a neural simulator for outfits. Our methodology yields skinned models with Pose Space Deformations through an implicit Physically Based Simulation using deep learning framework. This figure shows different neurally simulated outfits on different unseen body poses. Our results do not need collision-solving post-processing.}
    \label{fig:teaser}
\end{center}%
}]


\begin{abstract}
  We present a methodology to automatically obtain Pose Space Deformation (PSD) basis for rigged garments through deep learning. Classical approaches rely on Physically Based Simulations (PBS) to animate clothes. These are general solutions that, given a sufficiently fine-grained discretization of space and time, can achieve highly realistic results. However, they are computationally expensive and any scene modification prompts the need of re-simulation. Linear Blend Skinning (LBS) with PSD offers a lightweight alternative to PBS, though, it needs huge volumes of data to learn proper PSD. We propose using deep learning, formulated as an implicit PBS, to unsupervisedly learn realistic cloth Pose Space Deformations in a constrained scenario: dressed humans. Furthermore, we show it is possible to train these models in an amount of time comparable to a PBS of a few sequences. To the best of our knowledge, we are the first to propose a neural simulator for cloth. While deep-based approaches in the domain are becoming a trend, these are data-hungry models. Moreover, authors often propose complex formulations to better learn wrinkles from PBS data. Supervised learning leads to physically inconsistent predictions that require collision solving to be used. Also, dependency on PBS data limits the scalability of these solutions, while their formulation hinders its applicability and compatibility. By proposing an unsupervised methodology to learn PSD for LBS models (3D animation standard), we overcome both of these drawbacks. Results obtained show cloth-consistency in the animated garments and meaningful pose-dependant folds and wrinkles. Our solution is extremely efficient, handles multiple layers of cloth, allows unsupervised outfit resizing and can be easily applied to any custom 3D avatar.
\end{abstract}

\section{Introduction}
\label{sec:intro}

Animation of draped humans has been widely explored by the computer graphics community because of its wide range of potential applications: videogame, film industry, and nowadays, also in virtual and augmented reality VR/AR. We can split the different animation approaches based on their goal: performance or realism. On one hand, Physically Based Simulation (PBS) \cite{baraff1998large, liu2017quasi, provot1997collision, provot1995deformation, tang2013gpu, vassilev2001fast, zeller2005cloth} strategies discretize the space and time to apply basic physics laws. The realism obtained is closely related to how fine-grained is the discretization. At the same time, the computational cost greatly increase along with the level of realism. Moreover, simulation parameters need to be properly fine-tuned in order to obtain the desired results. Therefore, expert knowledge is necessary. On the other hand, Linear Blend Skinning (LBS) \cite{kavan2008geometric, kavan2005spherical, le2012smooth, Magnenat-thalmann88joint-dependentlocal, wang2007real, wang2002multi} and Pose Space Deformation (PSD) \cite{allen2002articulated, anguelov2005scape, lewis2000pose, loper2015smpl} techniques require a significantly lower amount of computational resources but compromise the realism of the animation. These strategies are suitable for low-computing environments or applications that demand real-time performance (portable devices and videogames). We close the gap between PBS and PSD by proposing an unsupervised approach to learn realistic PSD, and thus, maintaining their efficiency advantage against PBS.

Due to its recent success in complex 3D tasks~\cite{arsalan2017synthesizing, han2017deepsketch2face, madadi2020smplr, omran2018neural, qi2017pointnet, richardson20163d, socher2012convolutional}, we find deep learning as a promising approach to the garment animation problem. The research community has shown an increasing interest on draped 3D human animation through deep learning during the past few years \cite{alldieck2018video, alldieck2019tex2shape, bertiche2019cloth3d, bhatnagar2019multi, guan2012drape, lahner2018deepwrinkles, patel2020tailornet, santesteban2019learning}. Commonly, authors propose learning non-linear PSD models from big volumes of PBS data. Hence, these approaches also demand high computational resources to run the simulations. Another possibility for data gathering is through the use of 4D scans. While this solution allows capturing real data, it is necessary to build expensive and constrained setups. Furthermore, data obtained through scans need to be post-processed to be usable. Supervised deep learning based approaches not only depend on expensive data, but are also bounded by it. Plus, supervised training falsely assumes uniqueness of garment vertex locations, which, as we will show, hinders learning in practice.

In this paper we propose learning PSD for rigged garments leveraging deep learning framework and formulating the problem as an implicit PBS. By using PBS formulation, we force our models to predict consistent, low-energy configurations of the physical system that cloth and body represent. Doing this allows applying unsupervised training, removing the need of data gathering through expensive simulations or scans. Furthermore, we show that our proposed methodology can yield cloth-consistent PSD in a short amount of time (minutes). By eliminating the need of simulating hundreds or even thousands of sequences, we drastically reduce the time needed from garment design to model deployment. This increases the applicability and scalability of the methodology, broadening the scope of real life scenarios that will benefit from it. The final animated garments show cloth-consistency, pose-dependant wrinkles and temporal coherence for unseen pose sequences (see supplementary video). Fig. \ref{fig:teaser} shows some qualitative samples obtained with the methodology described in this paper. Our main contributions are:
\begin{itemize}
    \item \textbf{Unsupervised PSD Learning.} By enforcing physical consistency during the training of the model, we eliminate the dependency of PBS or scan data. As a consequence, this methodology can be applied to an arbitrary number of garments, body shapes and poses without the computational cost of obtaining data for them.
    \item \textbf{Efficient Training, Deployment and Compatibility.} Related deep based approaches in the current literature propose complex formulations to obtain realistic results. This hinders the training process and posterior deployment. Our proposed methodology yields blend shapes for LBS models, which is the standard for 3D animation, and therefore, it is automatically compatible with all graphic engines and benefits from the exhaustive optimization for these models. This greatly increases the applicability of the methodology.
    \item \textbf{Physical Consistency.} Learning based related works are unable to predict collision-free garments, and thus, require collision-solving post-processing to be used in real applications \cite{patel2020tailornet, vidaurre2020fully, jiang2020bcnet, gundogdu2019garnet, guan2012drape, santesteban2019learning}. This disables real-time performance, increases the required engineering effort to adapt such solutions and removes model differentiability, which hurts their applicability in research. Some of these works use instead collision-solving losses during training to alleviate the issue, but body interpenetration still appears. We show how this is related to supervised training. On the other hand, PBNS can generate collision-free predictions even under extreme unseen poses, effectively removing the need of post-processing. Physical consistency is not limited to collisions, but also to surface quality. Inspired by mass-spring models, we enforce edge and bending constraints in our predictions. This ensures outfits have no distorted edges ---which would generate texturing artifacts--- and smooth surfaces.
    \item \textbf{Cloth-to-cloth Interaction.} We are the first to propose a learning-based approach that is able to explicitly handle cloth-to-cloth interactions between different layers of cloth. Because of this, PBNS is the only current approach that can animate complete and complex outfits with multiple overlapping garments.
\end{itemize}

In Tab. \ref{tab:applicability} we compare the applicability of garment animation methodologies. We observe that current deep-learning based approaches require a large amount of computational resources to run, due to large models and/or the need of post-processing. For scenarios where resources are low or must be available for other tasks (videogames, portable devices, ...), these approaches cannot be applied. Furthermore, since formulations are complex and non-standard, the engineering cost of adapting them to real applications is prohibitive for small videogame or film studios or low-cost virtual try-ons. On the other hand, whenever computational resources are available, it is always preferable to run real-time GPU cloth simulation on high-end hardware. Our contribution, PBNS, is a methodology to generate per-outfit LBS models with PSD. These models have a memory footprint of a few megabytes and running times of over $14000$ samples per second (see Tab.~\ref{tab:performance}). Moreover, since training is unsupervised and takes a few minutes and output model formulation is the standard for 3D animation, the engineering cost for adaptation is minimal. Note that the use cases of our approach are the same as regular LBS models with PSD. PBNS increases the quality and realism of PSD. As aforementioned, these are the efficient alternative to PBS strategies for 3D animation. To the best of our knowledge, we are the first to propose a learning based methodology that can achieve real-time performance and realistic wrinkles in most environments (low-computing and/or portable devices), and thus, can be applied on real scenarios.

\begin{table}[]
\centering
\begin{tabular}{lccc}
\multicolumn{1}{c}{} & \multicolumn{1}{l}{PBS} & Deep learning & \multicolumn{1}{l}{PBNS} \\
\hline
Virtual Try-on & \checkmark & x & \checkmark \\
Videogames & x & x & \checkmark \\
Film & \checkmark & x & \checkmark \\
Virtual Reality & x & x & \checkmark \\
Portable devices & x & x & \checkmark \\
Garment resizing & x & \checkmark & \checkmark \\
\hline
\end{tabular}
\caption{Application possibilities for state-of-the-art in garment animation. Current deep-based approaches rely on the availability of computational resources for complex models. Whenever such resources are available, real-time GPU cloth simulation is always preferred. On the other hand, PBNS is a methodology to generate LBS models with PSD using deep learning framework. Models generated with PBNS take only a few megabytes in memory and can generate up to $14000$ samples per second.}
\label{tab:applicability}
\end{table}

\section{State-of-the-art}
\label{sec:sota}

In the computer graphics community, the garment animation problem has been tackled for decades. Although deep learning has shown significant progress during recent years, one of its main drawbacks in the case of garment animation is the scarcity of available data and the data-hungry nature of deep-based approaches.

\subsection{Computer Graphics} 

\textbf{PBS} (Physically Based Simulation) permits obtaining highly realistic cloth dynamics, usually relying on the \textit{spring-mass} model. The literature on this regard is exhaustive and mainly addresses the efficiency and robustness of the methodology. This is done through simplifications and specialization on constrained scenarios \cite{baraff1998large, provot1997collision, provot1995deformation, vassilev2001fast}. As another option, authors propose energy-based optimization approaches for an increase in stability and generalization to additional soft-bodies \cite{liu2017quasi}. Some works describe technical improvements to leverage the extra computational power that GPU parallelization yields \cite{tang2013gpu, zeller2005cloth, tang2018cloth}. Nonetheless, in spite of the increase on efficiency contributed by other works, achieving a high level of realism comes at a great computational expense. When such resources are not available or real-time performance is a must, PBS cannot be applied. To overcome this, \textbf{LBS} (Linear Blend Skinning) is used. LBS is the current standard for 3D animation in computer graphics. Objects motion is driven by an skeleton defined as a set of joints. Vertices of the mesh that represents the 3D object are attached to the joints by a set of blend weights. The transformation (rotation, translation and scaling) of each vertex is the weighted sum of the transformations of the joints with the aforementioned blend weights. Commonly, garments are attached to the same skeleton that controls the 3D body. We can also find an exhaustive research regarding LBS \cite{kavan2008geometric, kavan2005spherical, le2012smooth, Magnenat-thalmann88joint-dependentlocal, wang2007real, wang2002multi}. This approach allows real-time applications, even in low-computing environments, by sacrificing realism, specially on garment domain. Currently, we can find \textbf{hybrid} strategies in the industry. Tight parts of the outfits (e.g., t-shirt and trousers) are attached to the body skeleton while other apparel (e.g., capes and long coats) are simulated. This approach is widely used in the videogame industry, as realism is enhanced without an excessive increase on computational requirements.

\subsection{Learning-Based Approaches}

LBS models achieve real-time performance, nonetheless, linear transformations are usually not enough to capture the motion of soft-tissue objects such as cloth. Furthermore, LBS might suffer of skinning-related artifacts. \textbf{PSD} (Pose Space Deformation) aims to address these drawbacks by applying corrective deformations to LBS models before skinning \cite{lewis2000pose}. This helps reducing artifacts and also allows representation of high-frequency details that depend on the pose of the object. Hand-crafted PSD is intractable for complex models (such as the human body) and it is usually learnt from data. Authors have shown that PSD approaches are able to model the human body \cite{allen2002articulated, anguelov2005scape, loper2015smpl}. Deformations basis are obtained by linear decomposition of hundreds or thousands of 3D body scans. Following this fashion, for garments, Guant et al.\cite{guan2012drape} propose performing the same computation for synthetic garment data gathered through PBS. Later, L\"ahner et al.\cite{lahner2018deepwrinkles} extend the idea by computing the aforementioned linear decomposition against temporal feature arrays processed by a Recurrent Neural Network (RNN), achieving non-linearity w.r.t. the pose. Santesteban et al.\cite{santesteban2019learning} explicitly apply a non-linear mapping with a Multi-Layer Perceptron (MLP) for a single fixed garment. While these approaches achieve appealing results, each new garment or outfit requires repeating the simulation and learning process, thus hindering scalability and applicability. Authors commonly address this drawback by leveraging existing body models (like SMPL\cite{loper2015smpl}). Garments are encoded on top of the human body as subsets of displaced vertices \cite{alldieck2018video, alldieck2019tex2shape, bertiche2019cloth3d, bhatnagar2019multi, patel2020tailornet}. Following the idea of exploiting the human body model, Patel et al.\cite{patel2020tailornet} use subsets of body vertices as few different garments to later learn garment-specific models for high frequency cloth details. Bertiche et al.\cite{bertiche2019cloth3d} perform a similar encoding for thousands of different garments. This allows learning a continuous space for garment topology. Later they use each garment representation within this space to condition the pose-dependant vertex offsets. Using a body model to represent garments allows handling multiple types with a single model. Nonetheless, huge volumes of data are still needed to train these models. Furthermore, it has been proven that deep neural networks are biased to lower frequencies~\cite{rahaman2019spectral}, and as noted by Patel et al.\cite{patel2020tailornet}, this effect is more significant on cloth domain when training a single model to represent many different garment types. This means that in order to obtain high-resolution garment predictions, it is better to exhaustively simulate and train for individual garments. Our proposed methodology allows skipping the simulation step and efficiently learn, in few minutes, PSD for a given LBS model of a garment or outfit.

\section{Neural Cloth Simulation}
\label{sec:method}

Classical computer graphics approaches resort to manual template skinning and/or costly simulation, which compromise realism, performance or applicability. On the other hand, learning based approaches, non-deep and deep, propose a manual skinning followed by a data-driven method to compute Pose Space Deformations. Since simulated data is still necessary, a significant computational investment is required for each new garment, body shape or fabric. We propose learning garment-specific (or outfit-specific) PSD unsupervisedly by enforcing physical laws, addressing the main drawbacks of previous works in terms of data requirements.

\subsection{PBS Data and Physical Consistency}
\label{sec:phys_consistency}

As aforementioned, the current trend on this domain is supervised learning from PBS data. We will show how this is suboptimal for learning valid garment deformations. The mapping from pose-space to outfit-space is a multi-valued function. Different simulators, initial conditions, action speeds, timesteps and integrators, among other factors, will generate different valid outfit vertex locations for the same body pose and shape and outfit. Training or evaluating on PBS data falsely assumes that this mapping is single-valued. Samples with similar $\theta$ but significantly different $\mathbf{V}_{\theta}$ will hinder network performance during training and most likely converge to average vertex locations under a supervised loss. Moreover, a final user does not know the \textit{ground truth} and therefore cannot perceive the accuracy of the model w.r.t. PBS data, but the user can assess the physical consistency of the predictions (collision-free and cloth consistency). Minimizing Euclidean error w.r.t. \textit{ground truth} does not guarantee physical consistency, and therefore, the applicability of the obtained predictions in real life is limited. Recent works require post-processing to solve body penetrations \cite{patel2020tailornet, santesteban2019learning}. This partially defeats the purpose of using deep-learning, removes differentiability and real-time performance. Thus, standard PBS is always preferred against previous deep-based approaches. We propose a fully unsupervised training as an implicit physically based simulation to remove the need of post-processing. Because of this, and the extreme efficiency of our formulation, ours is the first approach that can be applied for real-time scenarios in most devices.

\subsection{Formulation}

Our goal is to obtain cloth-consistent PSD for a given garment or outfit rigged to the skeleton of an LBS body model, in order to animate cloth and body at once. The per-vertex formulation of skinning with PSD w.r.t. an articulated skeleton, represented as a set of joints $J\in\mathbb{R}^{K\times3}$, for a given template garment or outfit in rest pose $\mathbf{T}\in\mathbb{R}^{N\times3}$ is defined as:
\begin{equation}
    t'_i = \sum_k^K w_{k,i}G_k(\mathbf{\theta},J)(t_i + dt_i(\mathbf{\theta})),
    \label{eq:skinning_psd}
\end{equation}
where $w_{k,i}$ is the blend weight of vertex $i$ and joint $k$, $G_k$ is the linear transformation matrix corresponding to joint $k$, $\mathbf{\theta}$ is the skeleton pose in axis-angle representation, $t_i$ is the $i$-th garment vertex in rest pose and $dt_i$ is the pose space deformation corresponding to this vertex. We need to find a valid skinning as a set of blend weights $\mathcal{W}\in\mathbb{R}^{N\times K}$ and a PSD as a mapping $f : \mathbf{\theta} \rightarrow \{d\mathbf{T} \mid \mathbf{T}_{\mathbf{\theta}} = \mathbf{T} + d\mathbf{T}\}$ such that the output unposed garment $\mathbf{T}_{\mathbf{\theta}}\in\mathbb{R}^{N\times3}$ is properly aligned with the body (no collisions) and shows a realistic cloth-like behaviour after skinning. We propose using a neural network to approximate $f$. Note that our formulation is not dependant on the chosen body model, and the only requirement is to have a database of poses for the body.

\textbf{Blend Weights.} In the current literature we find authors that rely on the assumption that garments closely follow body motion \cite{santesteban2019learning, bertiche2019cloth3d, patel2020tailornet}. Results presented by these works prove it is a valid assumption that allows for a significant simplification of the problem. We rely on this to compute blend weights for our template garment. For each vertex $t_i\in\mathbf{T}$ we assign blend weights equal to those of the closest body vertex, with the body in rest pose (aligned with garment). Skirts break the assumption that cloth and skin are close to each other. For these kind of garments, we allow blend weights to be optimized along with network weights. We observed that doing this increases the model convergence speed. For other types of garments, we see no significant differences, except the computational overhead of optimizing blend weights along with the rest of the trainable parameters.

\textbf{PSD.} Skinning alone is not enough to properly model garments, as cloth behaviour is highly non-linear. For this reason, we formulate the model with PSD. Classical computer graphics approaches rely on linear decomposition from training samples to obtain a PSD matrix like $\mathbf{D}\in\mathbb{R}^{\lvert\mathbf{\theta}\lvert\times N\times3}$, where $\lvert\mathbf{\theta}\lvert$ is the dimensionality of the pose array and $N$ is the number of vertices of the 3D mesh to animate. We propose to first obtain a high level embedding of the pose array $\mathbf{\theta}$ through a neural network as $\mathbf{X} = f_\mathbf{X}(\mathbf{\theta})$ and use this embedding to obtain the final deformations with a PSD matrix as $\mathbf{D}\in\mathbb{R}^{\lvert\mathbf{X}\lvert\times N\times3}$. This allows modelling any non-linear mapping from $\mathbf{\theta}$ to $dT$ thanks to the universal approximation properties of neural networks and also to control matrix $\mathbf{D}$ size and capacity. Learning of both $f_\mathbf{X}$ and $\mathbf{D}$ is performed following a deep learning framework, where variables are optimized by minimizing a loss function with batches of different input pose samples.

\subsection{Architecture}
\label{sec:architecture}

\begin{figure*}
    \centering
    \includegraphics[width=\textwidth]{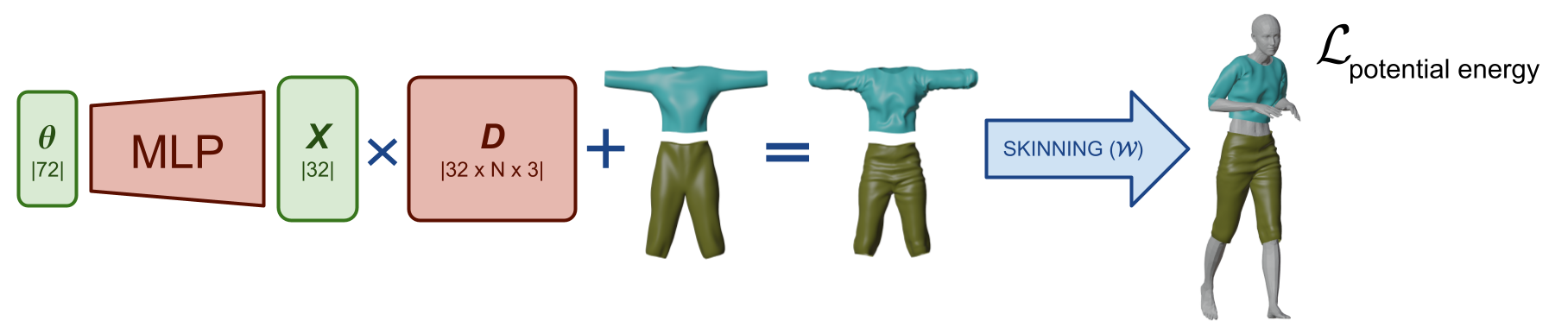}
    \caption{Methodology overview. Pose parameters are fed to a $4$-layer Multi-Layer Perceptron with $32$ dimensions in each layer. The output high-level pose embedding is multiplied with a PSD matrix $\mathbf{D}$ to obtain deformations for the garment/outfit. As standard, deformations are applied in rest pose garments and skinned along with the body (note that garment blend weights are assigned by proximity w.r.t. the body in rest pose). Finally, we train by optimizing the potential energy of the output physical system represented by the body and the cloth. The parts of the model in red correspond to the trainable parameters.}
    \label{fig:pipeline}
\end{figure*}

We design $f_\mathbf{X}$ as a Multi-Layer Perceptron (MLP). More specifically, $4$ fully connected layers with a dimensionality of $32$ and ReLU activation function. Then, to obtain the posed outfit $\mathbf{V}_\mathbf{\theta}$ for a given $\mathbf{\theta}$:

\begin{equation}
    \mathbf{V}_\mathbf{\theta} = W(\mathbf{T} + f_\mathbf{X}(\mathbf{\theta})\cdot\mathbf{D}, \mathbf{\theta}, \mathcal{W})
\end{equation}

Where $W(\cdot, \mathbf{\theta}, \mathcal{W})$ is the skinning function for pose $\mathbf{\theta}$ and blend weights $\mathcal{W}$, and the product $f_\mathbf{X}(\mathbf{\theta})\cdot\mathbf{D}$ is computed as $\sum_i^{\lvert\mathbf{X}\lvert} f_\mathbf{X}(\mathbf{\theta})_i\mathbf{D_i}$. Under this formulation, we can approximate any non-linear mapping from $\mathbf{\theta}$ to $\mathbf{V}_\mathbf{\theta}$ while keeping compatibility with current graphic engines. Fig. \ref{fig:pipeline} shows an overview of the proposed methodology. The input of the model is the pose array $\mathbf{\theta}$, which is processed through the aforementioned MLP to yield a high-level pose embedding $\mathbf{X}$. This embedding $\mathbf{X}$ is multiplied with the PSD matrix $\mathbf{D}$ to obtain garment vertex deformations. Both, the MLP and the matrix $\mathbf{D}$ are learnt through training (and blend weights $\mathcal{W}$ are optimized from their initial proximity-based values for outfits with skirt). Finally, the deformed garment (or outfit) is skinned along the body according to $\mathbf{\theta}$ and blend weights $\mathcal{W}$. For the rest of the paper, for clarity reasons, we consider the PSD matrix $\mathbf{D}$ to be part of the network.

\subsection{Training}
\label{sec:training}

Just as physical systems are implicitly \textit{optimized} by acting forces $\mathbf{F} = -\mathbf{\nabla}U$ (where $U$ is the potential energy of the system), we train our neural network under a loss defined as a potential energy. This way, our model will learn to predict consistent, low-energy stable configurations. We define our global loss (or potential energy) as:

\begin{equation}
    \mathcal{L} = \mathcal{L}_{cloth} + \mathcal{L}_{collision} + \mathcal{L}_{gravity} 
    + \mathcal{L}_{pin},
    \label{eq:energy}
\end{equation}
where $\mathcal{L}_{cloth}$ corresponds to the elastic potential energy of the garment, and guides the model to predict cloth-consistent meshes. The term $\mathcal{L}_{collision}$ formulates body penetrations as a potential energy, thus its gradients will push cloth vertices to valid locations. Finally, $\mathcal{L}_{gravity}$ is the gravitational potential energy, which will minimize vertices height within the constraints set by the other loss terms. Additionally, we define $\mathcal{L}_{pin}$ to regularize deformations of chosen vertices (inspired by PBS).

\textbf{Cloth consistency.} The first term of our loss is related to the cloth consistency of the predictions. That is, we want the output meshes to fulfill certain properties we find in cloth. Classical computer graphics approaches usually rely on the mass-spring model to simulate cloth. We extend this idea to deep learning by designing a cloth loss term as:
\begin{equation}
    \mathcal{L}_{cloth} = \lambda_e\mathcal{L}_{edge} + \lambda_b\mathcal{L}_{bend} = \lambda_e\left\Vert E - E_{\mathbf{T}}\right\Vert^2 + \lambda_b\Delta(\mathbf{N})^2,
    \label{eq:cloth}
\end{equation}
where $E\in\mathbb{R}^{N_E}$ is the predicted edge lengths, $E_{\mathbf{T}}\in\mathbb{R}^{N_E}$ is the edge lengths on the rest garment $\mathbf{T}$ (with $N_E$ as the number of edges in the mesh), $\mathbf{N}\in\mathbb{R}^{N_F\times3}$ is the face normals (for $N_F$ triangular faces), $\Delta(\cdot)$ is the Laplace-Beltrami operator and $\lambda_e$ and $\lambda_b$ are balancing factors. On the one hand, the term $\mathcal{L}_{edge}$ ensures that cloth is not excessively stretched or compressed. It is formulated as the potential elastic energy of the system, such that its gradients act as forces. On the other hand, $\mathcal{L}_{bend}$ enforces locally smooth surfaces by penalizing differences between neighbouring face normals (hinge-like forces). Note that the latter is computed taking into account face connectivity, not vertex.

\textbf{Collisions.} Next, the model needs to handle collisions with the body model. To do so, we design the following loss:
\begin{equation}
    \mathcal{L}_{collision} = \lambda_c\sum_{(i,j)\in A} min(\mathbf{d}_{j,i}\cdot \mathbf{n}_j - \epsilon, 0)^2,
    \label{eq:collision}
\end{equation}
where $A$ represents the set of correspondences $(i,j)$ between predicted outfit and body, respectively, through nearest neighbour, $\mathbf{d}_{j,i}$ is the vector going from the $j$-th vertex of the body to the $i$-th vertex of the outfit, $\mathbf{n}_j$ is the $j$-th vertex normal of the body, $\epsilon$ is a small positive threshold to increase robustness and $\lambda_c$ is a balancing weight ($\epsilon = 4mm$ and $\lambda_c = 25$ in our experiments). This loss is crucial to obtain valid predictions and its gradients will push outfit vertices outside the body. It is designed under the assumption that cloth closely follows the skin, which we can safely assume given that initial skinning blend weights are assigned by proximity. While this is a naive implementation of a collision loss, it works well in practice and similar L1 formulations have already been used in deep-based approaches \cite{tiwari2020sizer, jiang2020bcnet, gundogdu2019garnet}. We opted for a quadratic term to enhance generalization and stability, plus, it helps achieving a balance w.r.t. the other loss terms. Note that, as with PBS, invalid bodies (self-collided) might corrupt the results.

\textbf{Cloth-to-cloth.} To be able to model whole outfits, it is necessary to explicitly handle cloth-to-cloth interactions. To this end, we define a layer order for each garment of a given outfit, from inner to outer. Then, we iteratively apply Eq.~\ref{eq:collision} to each layer, computing correspondences $A$ using body and previous layers. This will simulate repelling forces for both vertices $(i,j)\in A$ whenever correspondences connect two different cloth layers. To the best of our knowledge, we are the first to explicitly tackle cloth-to-cloth interaction for learning based approaches.

\textbf{Gravity.} We include an additional term to enforce more realistic garment predictions. This term models the effect of the gravity. From classical mechanics, we know that potential gravitational energy is $U = m\cdot g\cdot h$, where $m$ is the mass of the object, $g$ is the gravity and $h$ is the height of the object. Since $m$ and $g$ are constant, we can understand this loss as $\mathcal{L}_{gravity} = k {\mathbf{V}_{\mathbf{\theta}}}_z$. In other words, we are minimizing the $Z$ coordinate (vertical axis) of each vertex of the predicted garments.

\textbf{Pinning.} For some garments we want certain vertices not to move around. For example, lower body garments might fall down as training progresses due to the gravity loss. We want to restrict waist vertices deformation such that it remains attached to its original position. The concept of pinning appears in most cloth simulators. To this end, we implement an L$2$ regularization loss on the deformations $dt$ of each vertex defined as pinned down. That is, $\mathcal{L}_{pin} = \lambda_{pin}\sum_ib_idt_i^2$ where $b_i = 1$ if vertex $i$ is \textit{pinned}, else $b_i = 0$. Note that a hard constraint would most likely produce collisions against the body, so vertices need to be able to move slightly. Then, we include it in the loss as an extra term with a balancing weight $\lambda_{pin} = 10$.

By formulating both $\mathcal{L}_{cloth}$ and $\mathcal{L}_{gravity}$ as physical magnitudes, their corresponding loss balancing weights are directly related to the properties of the fabric we want to simulate: Young's modulus for the elasticity and its mass for gravity. This provides of explainability to the approach. The rest of the losses, as with classical computer graphics PBS, are simplifications of the underlying physics.

\section{Experiments}
\label{sec:experiments}

In this section we will first describe the process to apply the methodology explained in Sec.~\ref{sec:method}. Then, we define the data and setup for the experimental part.

\subsection{Body model}

SMPL \cite{loper2015smpl} is the current standard in the literature for human analysis and garment animation. This model is an LBS with PSD obtained through thousands of accurate 3D scans of different subjects. Its underlying skeleton is defined as a set of $K = 24$ joints. Public pose databases are available for this model (AMASS \cite{mahmood2019amass}). We then choose SMPL for the experimental part because both model and pose data are available to the public. Nonetheless, the methodology described in this paper is compatible with any 3D model rigged to an skeleton. SMPL also allows generating different body shapes through blend shapes. During neural simulation, body shape is fixed (just as with PBS, where, in general, we do not want the body shape to change during simulation).

\subsection{Template outfit}

Once a body model is selected, a garment or outfit is designed for the body in rest pose, with an approximate resolution of $1$cm in our experiments. We smooth templates as much as we can before neural simulation. This will ensure that high frequency details and deformations are indeed generated by the model from the pose. Then, initial blend weights for the cloth are obtained by proximity to the body.

\subsection{Pose database}
\label{sec:poses}

Neural simulation requires a database of valid poses for the selected body model. We define, as valid poses, those that do not produce self-collisions when applied to the 3D body model. As with regular PBS, neurally simulating cloth over bodies with self-collisions will generate inconsistent repelling forces and might corrupt the results. For SMPL, we have $\lvert\mathbf{\theta}\lvert = 3K = 72$. We choose CMU MoCap pose sequences. This dataset contains $2667$ pose sequences of different length, performed by different subjects. It totals around $4.3$M individual poses. We split the database into train and test per subject, thus ensuring no subject or sequence is repeated in both sets, as $85/15\%$. Then, to ensure pose balance, we randomly sample $N = 3000$ poses from the training set, such that not any pair of poses have a distance $d < 0.5$, with $d = max(\lvert\mathbf{\theta}_i - \mathbf{\theta}_j\lvert)$. Thus, for any two poses, there is at least one parameter with a difference equal or bigger than $0.5$ radians (we omit global orientation for this sampling). Later, we split the $N = 3000$ samples into training and validation set as $85/15\%$ ($2550$ training poses and $450$ validation poses).

\section{Results}
\label{sec:results}

In this section we present the results obtained through experiments. First, a justification of the chosen architecture for the model. Then, a comparison against standard supervised learning. Later, we display and discuss qualitative results. Next, we show some interesting properties of neural simulation. Following, a comparison against the state-of-the-art. As it is standard in deep learning, all the presented results, quantitative or qualitative, correspond to unseen test sequences. This is also true for the supplementary video. Finally, we illustrate more possibilities for neural simulation: outfit resizing and custom avatar enhancement.

\subsection{Multi-Layer Perceptron}

\begin{figure*}
    \centering
    \includegraphics[width=\textwidth]{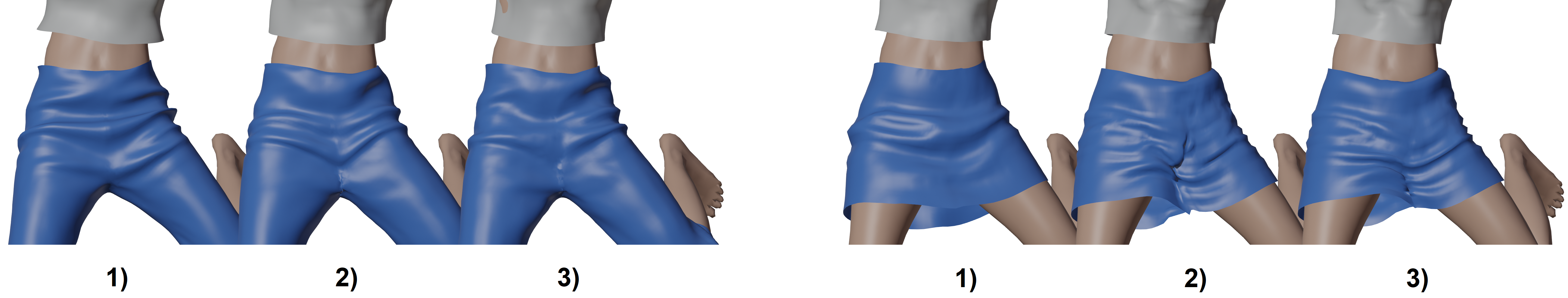}
    \caption{Architecture ablation study. We perform three experiments to assess the effect of the proposed MLP: 1) PBNS as described, 2) $\theta$ as input with no MLP and 3) rotation matrices $\mathbf{R}_{\theta}$ as input with no MLP. For experiment 2 we observe unrealistic \textit{V}-shaped wrinkles around the hip. Also, for both models with no MLP, we observe artifacts where both legs merge. This effect is more evident on skirts. The merging point of left and right side of the skirt presents an important artifact for experiments with no MLP, despise being trained during hundreds of epochs (as opposed to $15$ epochs for PBNS with MLP).}
    \label{fig:ablation}
\end{figure*}

In Sec.~\ref{sec:architecture} we discussed the motivation for the MLP. We found it is possible to apply this methodology without an MLP by using pose $\theta$ to linearly combine the blend shapes within matrix $\mathbf{D}$ ($f_\mathbf{X}(\theta) = \theta$ in Eq.~\ref{eq:skinning_psd}). Nonetheless, we observe an MLP presents important advantages. First, the size of matrix $\mathbf{D}$ is several orders of magnitude larger than the proposed MLP. Thus, controlling the size of this matrix allows for more efficient models. Without an MLP, the dimensionality of $\mathbf{D}$ is fixed to $\mathbb{R}^{\|\theta\|\times N\times3}$ ($\|\theta\| = 72$ blend shapes for SMPL as body model), and each of these blend shapes would be associated to a single $\theta$ parameter. This is clearly sub-optimal as some of these blend shapes would be irrelevant. Also, blend shapes would be linearly tied to their corresponding pose parameters. This is also sub-optimal since the mapping from axis-angle space to Euclidean space is non-linear. Moreover, cloth deformations are likely to be non-linear w.r.t. body pose. With an MLP, the model learns more meaningful blend shapes that are combined non-linearly w.r.t. $\theta$. Empirically, we also find the training to be faster and more stable with an MLP. We perform three different experiments to study the effects of an MLP: 1) PBNS as proposed in this paper, 2) without MLP w.r.t. $\theta$ and 3) without MLP w.r.t. the rotation matrices $\mathbf{R}_{\theta}$ generated from $\theta$ (to alleviate non-linear relations between axis-angle space and Euclidean space). Fig.~\ref{fig:ablation} presents the results of these experiments. Each sample has a number that corresponds to one of the experiments explained. We see how experiment $2$ presents unrealistic \textit{V}-shaped wrinkles around the hip (left samples). Experiment number $3$ does not show this. It means this is the result of linearly approximating a non-linear relation between axis-angle and Euclidean space. On top of this, we also observe artifacts on the region where both legs merge for experiments with no MLP. This effect is more evident on skirts (right samples). These artifacts are present even after hundreds of training epochs (note that experiment $1$ is trained for just $15$ epochs).

\subsection{Supervised learning and quantitative evaluation}
\label{sec:supervised}

\begin{figure*}[ht]
    \centering
    \includegraphics[width=\textwidth]{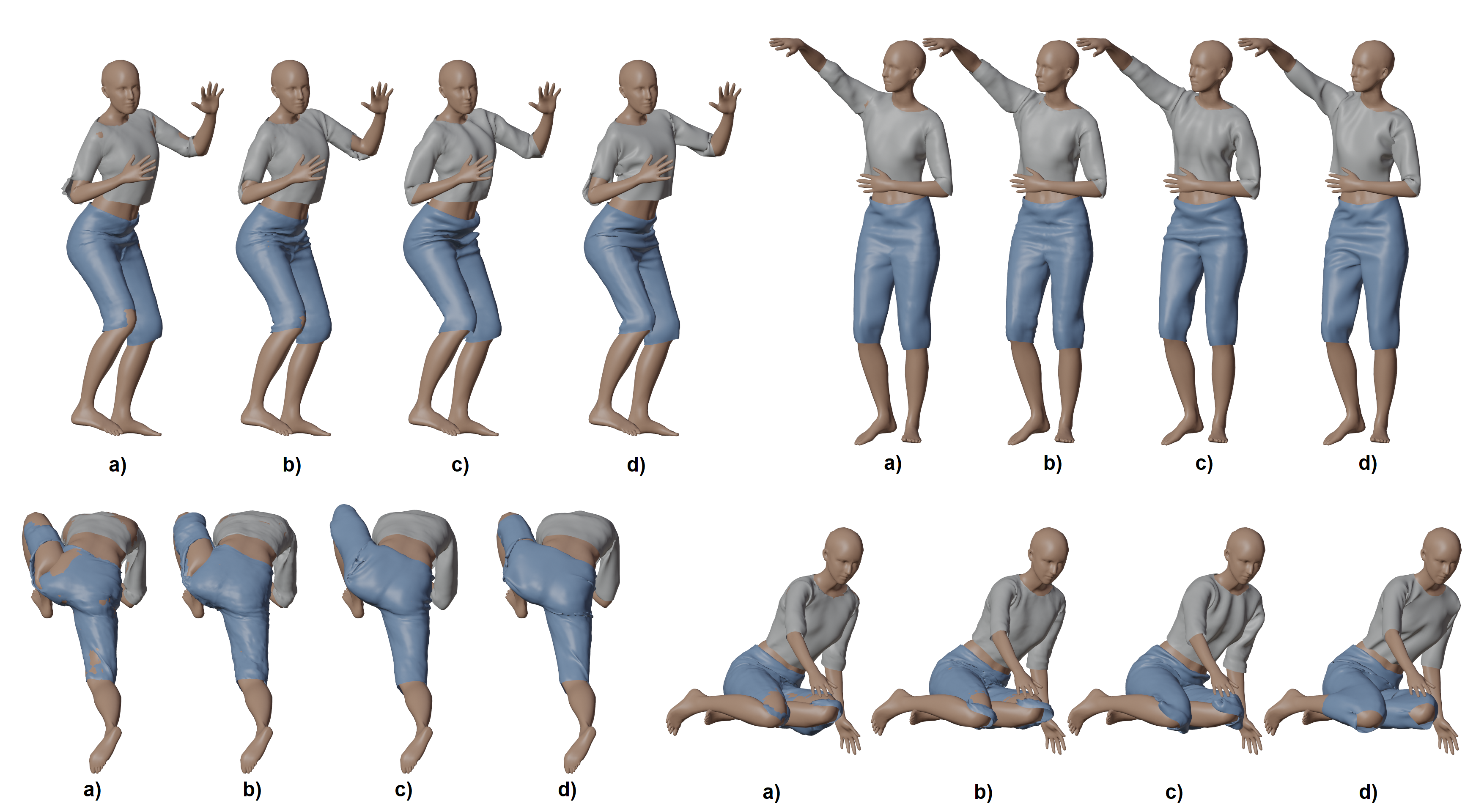}
    \caption{Supervision vs PBNS. We compare results obtained with PBNS against a purely supervised approach and a hybrid approach (L2 plus Eq.~\ref{eq:energy}). The same architecture and the same $N = 3000$ poses ($2550$ for training and $450$ for validation) are used in these experiments. We additionally compare against PBS data used for supervised training. We show: a) L2 only, b) hybrid, c) PBNS and d) PBS data. The supervised approach has the lowest Euclidean error w.r.t. PBS data, but also lowest number of wrinkles and highest number of collisions. The hybrid approach has more visible wrinkles and fewer collisions, but generalizes poorly to extreme poses. PBNS has the highest Euclidean error, but realistic pose-dependant wrinkles and shows no collisions, even under extreme poses. We also observe how PBS is prone to failures when body presents self-collisions (upper-left sample right elbow and lower-right sample legs). These failures might be transferred to predictions if trained supervisedly. PBNS is more robust to collisions than PBS.}
    \label{fig:vs_supervised}
\end{figure*}

\begin{table}[t!]
\centering
\begin{tabular}{lcccc}
\hline
Method & Error (mm) & Edge (mm) & Collision & Time\\
\hline
L2 & 7.59 & 0.78 & 3.15\% & $\sim30$h \\
Hybrid & 8.21 & 0.74 & 1.08\% & $\sim30$h \\
PBNS & 15.52 & 0.66 & 0.45\% & $\sim15$m \\
\hline
\end{tabular}
\caption{Quantitative comparison of supervised baseline vs. PBNS. Each row represents an experiment. All experiments were performed with the same architecture. First row shows a purely supervised experiment, with an L2 loss w.r.t. PBS data. The second row combines L2 loss with Eq.\ref{eq:energy}. Finally, the last row corresponds to PBNS as described in this paper. For each experiment we report Euclidean error against PBS data, average edge compression/elongation w.r.t. edge rest lengths, the ratio of collided cloth vertices (within human body) and the time it takes to obtain a trained model. We observe how supervision compromises physical consistency (edge distortion and higher collision ratio). Since supervision requires PBS data, simulation time has to be taken into account. On the other hand, while Euclidean error is higher, PBNS can generate cloth-consistent and almost collision-free predictions in a very short amount of time.}
\label{tab:supervised}
\end{table}

To compare our approach against supervised learning and provide of a quantitative metric, we compute PBS data for the $N = 3000$ poses described in Sec.~\ref{sec:poses}. Once these data is obtained, we perform three different experiments (shown in Tab.~\ref{tab:supervised}). In the first row, we train the described architecture supervisedly with an standard L2 loss on the predictions w.r.t. PBS data. In the second row, we train the same model with L2 loss combined with Eq.~\ref{eq:energy}. In the last row, we evaluate our unsupervised training results against the PBS data. We complement this table with a qualitative comparison shown in Fig.~\ref{fig:vs_supervised}. From left to right: a) L2 only, b) hybrid, c) PBNS and d) PBS data. As can be seen, the supervised approach is able to minimize Euclidean error w.r.t. PBS data. Nonetheless, in order to do so, it compromises physical consistency. The supervised approach is prone to collisions, which makes predictions unusable in real applications. We also observe a minimal amount of wrinkles. The hybrid approach shows a lower number of collided vertices and slightly more visible wrinkles. Nonetheless, as can be seen, it generalizes poorly to extreme poses (lower row samples). Also, in the upper-left sample, we see a failure in the left elbow that is not present on PBS data. Thus, combining L2 loss with physical consistency has an unpredictable behaviour. Then, we see how PBNS can generate cloth-consistent and collision-free predictions, even under extreme poses. Finally, we also show PBS data in the figure. As can be seen, PBS might fail for poses that present body self-collisions (upper-left sample right elbow and lower-right sample legs). PBS failures might be transferred to predictions through supervision. Note how PBNS is more robust to failures than classic PBS. Finally, we also compare the amount of time devoted to obtain each model in the table. For the supervised and hybrid approaches, we need PBS data, which has a large computational cost. In the time needed to obtain a single animated outfit through supervised approaches, PBNS can generate over a hundred of different models for different outfits. Overall, we have shown how unsupervised training is not only more efficient (no need to generate PBS data), but it also qualitatively outperforms supervised approaches. Furthermore, it shows higher robustness against simulation failures than traditional PBS. We have also shown how Euclidean error is misleading (lower does not mean better).

\subsection{Qualitative}

\begin{figure*}
    \centering
    \includegraphics[width=.9\textwidth]{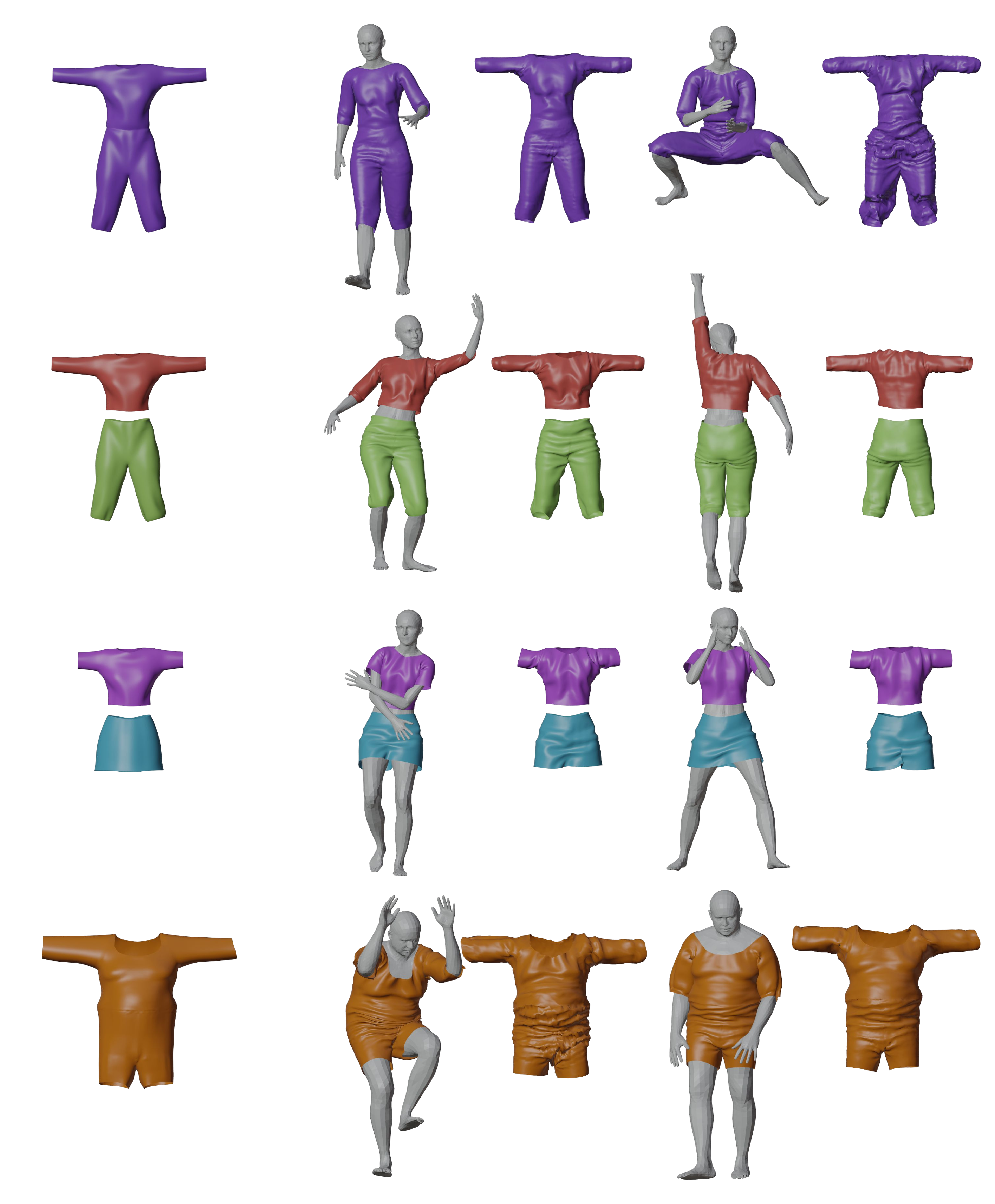}
    \caption{Qualitative results. Leftmost, we depict the template outfit of each row. Note how templates are smoothed as much as possible. Then, for each row we visualize the output for two unseen test poses. For each sample, we show the output dressed human (left) along with its corresponding deformed outfit in rest pose (right). As it can be seen, the learnt Pose Space Deformations generate folds and wrinkles to fulfill the energy balance requirement of the physical system. The last row depicts results of simulation with a different body shape (and its appropriately aligned template outfit) to show generalization to different bodies. All of the shown samples were obtained after just a few minutes of training.}
    \label{fig:psd}
\end{figure*}

\begin{figure*}
    \centering
    \includegraphics[width=\textwidth]{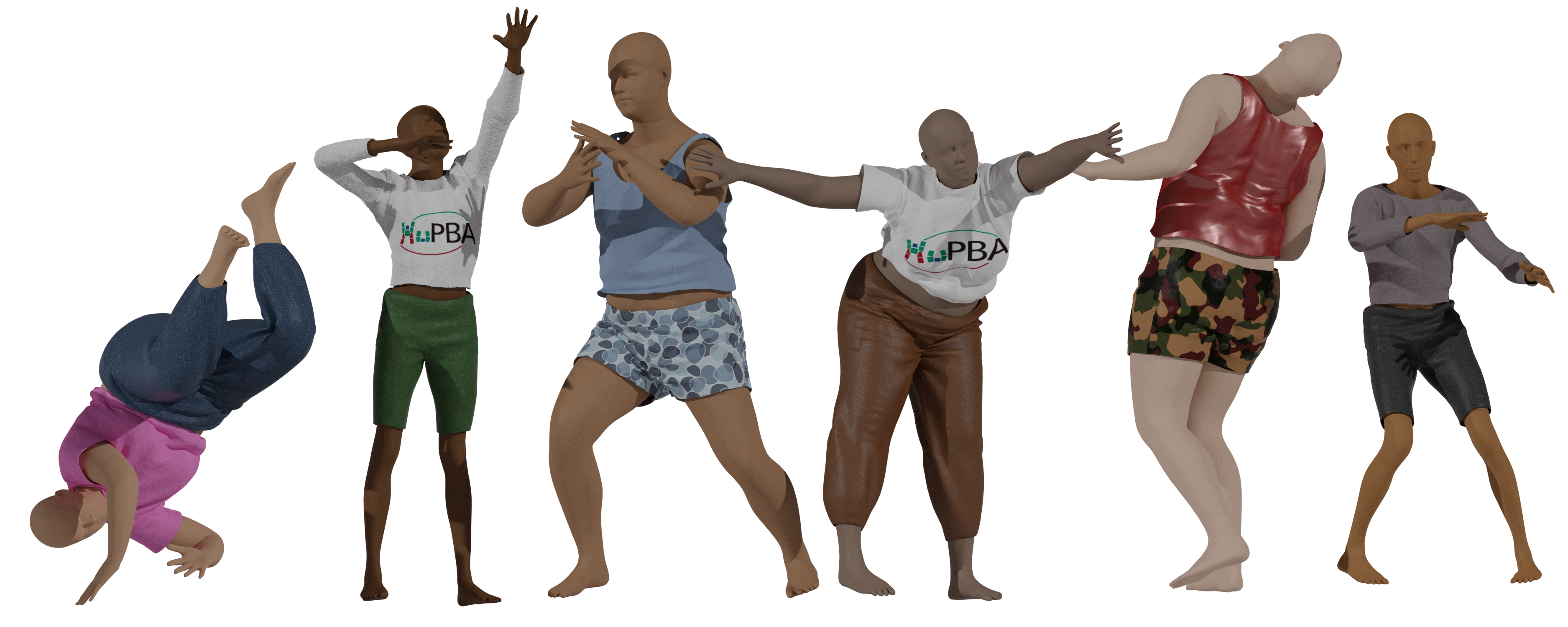}
    \caption{Textured qualitative samples. This rendering gives an approximate idea to the final application-level looks that our methodology yields. These results correspond to neural simulations of $3$ additional outfits on top of $3$ different bodies, further showing the generalization capabilities of our methodology.}
    \label{fig:textured}
\end{figure*}

For a qualitative evaluation of the results, we refer first to Fig. \ref{fig:teaser}. In this image we show a few samples with different pose, outfit and, one of them, different body. As it can be seen, our learnt PSD can generate appropriate wrinkles around bent joints in a realistic manner to fulfill the energy balance requirements imposed by our loss during training. Then, for a more in-depth analysis, we refer to Fig. \ref{fig:psd}. Here we show, on one hand, the template outfit of each sample. Templates are smoothed as much as possible. Later, for each template, the output for two extra unseen poses are shown (different from Fig. \ref{fig:teaser}). For each of these samples, the final rigged draped human is visualized (left) along with the unposed deformed template garment $\mathbf{T}_{\mathbf{\theta}}$ (right). Templates show deformations to satisfy the energy constraints which would not be possible with skinning alone. From the first row, we can notice how big deformations are due to collisions for extreme poses. Also, while deformed template can look noisy, it looks realistic after skinning. On the second row we can see deformations on the back of the outfit. Nonetheless, as human bodies usually bend forward, wrinkles in the front are more evident. The third row shows samples with a skirt. Note how in the second sample, the skirt deformations need to correct rotations due to leg movements (deformed template has a discontinuity). For outfits with skirt, we allow optimization of blend weights, along with the rest of the network, to alleviate the correction required due to leg motion. In spite of this, the effect on the template is not fully mitigated. Finally, the last row shows results obtained with a different body shape. As aforementioned, the methodology presented in this paper is compatible with any 3D model rigged to an skeleton. On Fig. \ref{fig:textured} we rendered more qualitative results for different bodies and outfits. We refer the reader to the supplementary video for more qualitative results. All of these visualizations correspond to models trained during just a few minutes without any post-processing.

\subsection{Multiple Layers and Controllable Parameters}

\begin{figure*}
    \centering
    \includegraphics[width=.9\textwidth]{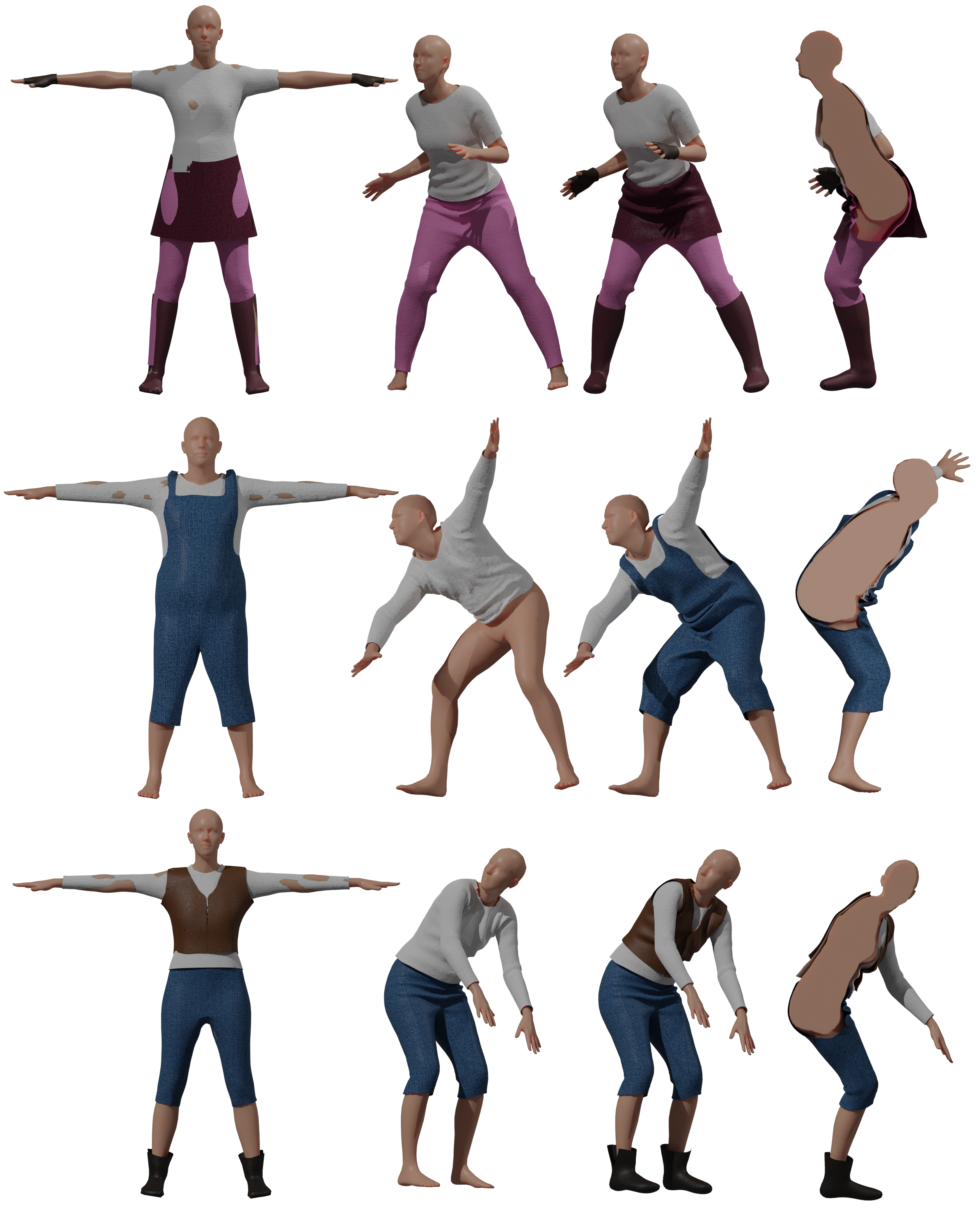}
    \caption{PBNS results on outfits with overlapping layers of cloth. Each row shows a sample. From left to right: template outfit, posed sample without outermost layer, posed sample with whole outfit and cross section. The proposed collision loss can handle multiple layers of cloth, even when the template outfit presents inter-penetrations. As we can observe on the cross section, PBNS is able to generate consistent predictions with minimal layer spacing. Also note how it is possible to neurally simulate different fabrics --different wrinkle count and size-- within the same outfit by assigning per-vertex weights to cloth related losses.}
    \label{fig:layers}
\end{figure*}

The proposed collision loss can be extended to deal with multiple layers of cloth (see Sec.~\ref{sec:training}). Fig.~\ref{fig:layers} contains the results obtained for some outfits that present cloth-to-cloth interaction. For each sample, from left to right: outfit in rest pose, posed outfit without outer layer, whole posed outfit and cross section. First, we see how templates in rest pose show collisions, against the body and cloth-to-cloth. The collision loss term is able to recover all of this inter-penetrations. Note that not even standard PBS is able to recover inter-penetrations between open meshes (cloth-to-cloth). Thus, again, PBNS appears to be more robust than PBS against collisions. Nonetheless, PBS is a more general solution. This property can be convenient for 3D artists, as it allows faster outfit design without hurting the final results. These results also show PBNS can handle cloth-to-cloth interactions accurately with minimal layer spacing. In the first sample, near the waist, we see up to three overlapping layers of cloth correctly sorted out --four layers if we consider the body. Additionally, different layers of cloth show different wrinkle count and size. This is due to the controllability of neural simulation parameters. Just like classic PBS, it is possible to simulate different fabrics by assigning per-vertex weights for $\mathcal{L}_{edge}$ and $\mathcal{L}_{bend}$. Higher weights will produce fewer, larger wrinkles (outer layers in the figure). Finally, the figure also shows the possibility to neurally simulate complements like gloves and boots. Rigid objects (boots) can be included into the neural simulation through parameter controllability (high weights). We are the first to propose a learning based methodology able to deal with cloth-to-cloth interactions, to allow result controllability by tuning parameters related to fabric physical properties --enhancing explainability-- and defining a common framework for garments and complements.

\subsection{Comparison}

\begin{figure*}
    \centering
    \includegraphics[width=\textwidth]{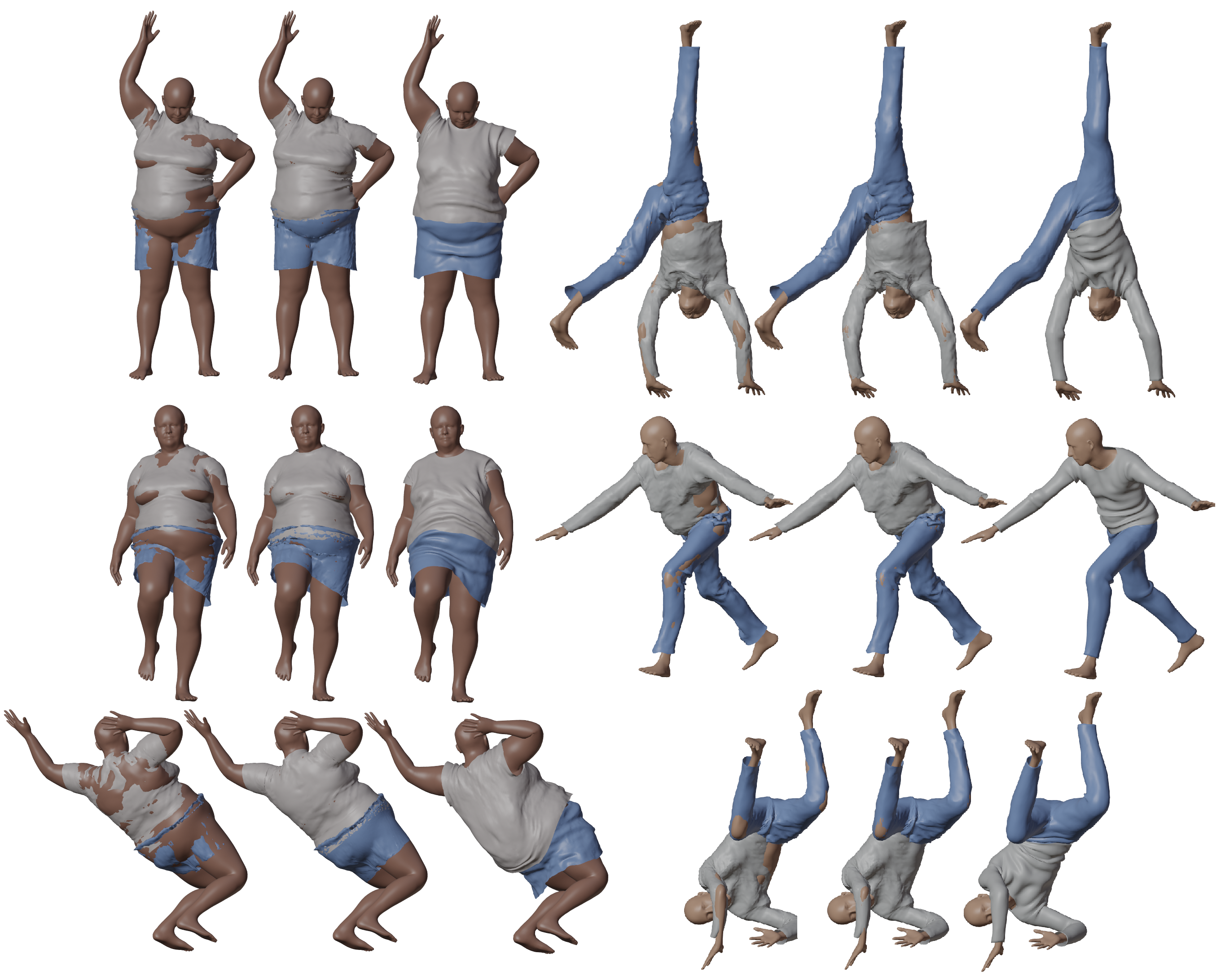}
    \caption{PBNS vs. TailorNet. We show two different bodies and outfits in three different poses each. For each pose, from left to right: TailorNet raw predictions, TailorNet after post-processing and PBNS. As observed, TailorNet heavily relies on post-processing, hurting performance and removing model differentiability. TailorNet predictions are noisy, show a bias towards body geometry and is unable to model cloth-to-cloth interactions. On the other hand, PBNS predictions show physically cloth-consistent predictions without body or cloth-to-cloth interpenetrations (raw predictions).}
    \label{fig:vs_tailornet}
\end{figure*}

The current reference in the garment animation domain within the context of deep learning is TailorNet \cite{patel2020tailornet}. We qualitatively compare results obtained with PBNS against TailorNet predictions. Note that TailorNet also addresses garment edition and resizing along animation. We choose two different body shapes and outfits and compare predictions obtained with both models. Authors of TailorNet post-process their predictions to solve collisions. For the sake of the comparison, since PBNS does not require post-processing, we show TailorNet with and without post-processing. Note that post-processing hugely increases computational time and removes model differentiability. Fig.~\ref{fig:vs_tailornet} shows the results obtained. For each sample we show, from left to right: TailorNet raw predictions, TailorNet post-processed and PBNS. As can be seen, TailorNet predictions heavily rely on post-processing. TailorNet models individual garments independently, and it is thus unable to handle cloth-to-cloth interactions. On the other hand, PBNS can almost guarantee collision-free predictions even under extreme poses for cloth-to-body and cloth-to-cloth interactions. We can see that PBNS predictions are less noisy and better resembles cloth. Since TailorNet encodes garments as body offsets, body geometry is transferred to predictions for unseen body shapes (middle right and bottom right samples). We observe TailorNet quality diminishes as we move outside their training pose and shape distribution. Moreover, while both approaches are static (no dynamics), we observe TailorNet is unable to achieve temporal consistency, while PBNS does (see supplementary video). In conclusion, while TailorNet compromises physical consistency and generalization to learn wrinkles, PBNS is able to generate pose-dependant wrinkles for completely unseen poses by imposing physical consistency. Additionally, PBNS complexity and computational time is several orders of magnitudes lower than TailorNet. TailorNet model size is in the order of gigabytes, while PBNS requires a few megabytes. Regarding computational time, for TailorNet we obtain around $3-5$FPS without post-processing and $0.6-0.9$FPS with post-processing, while PBNS can easily achieve hundreds or even thousands of executions per second (see Tab.~\ref{tab:performance}).

\subsection{Garment Resizing}

\begin{figure}
    \centering
    \includegraphics[width=\columnwidth]{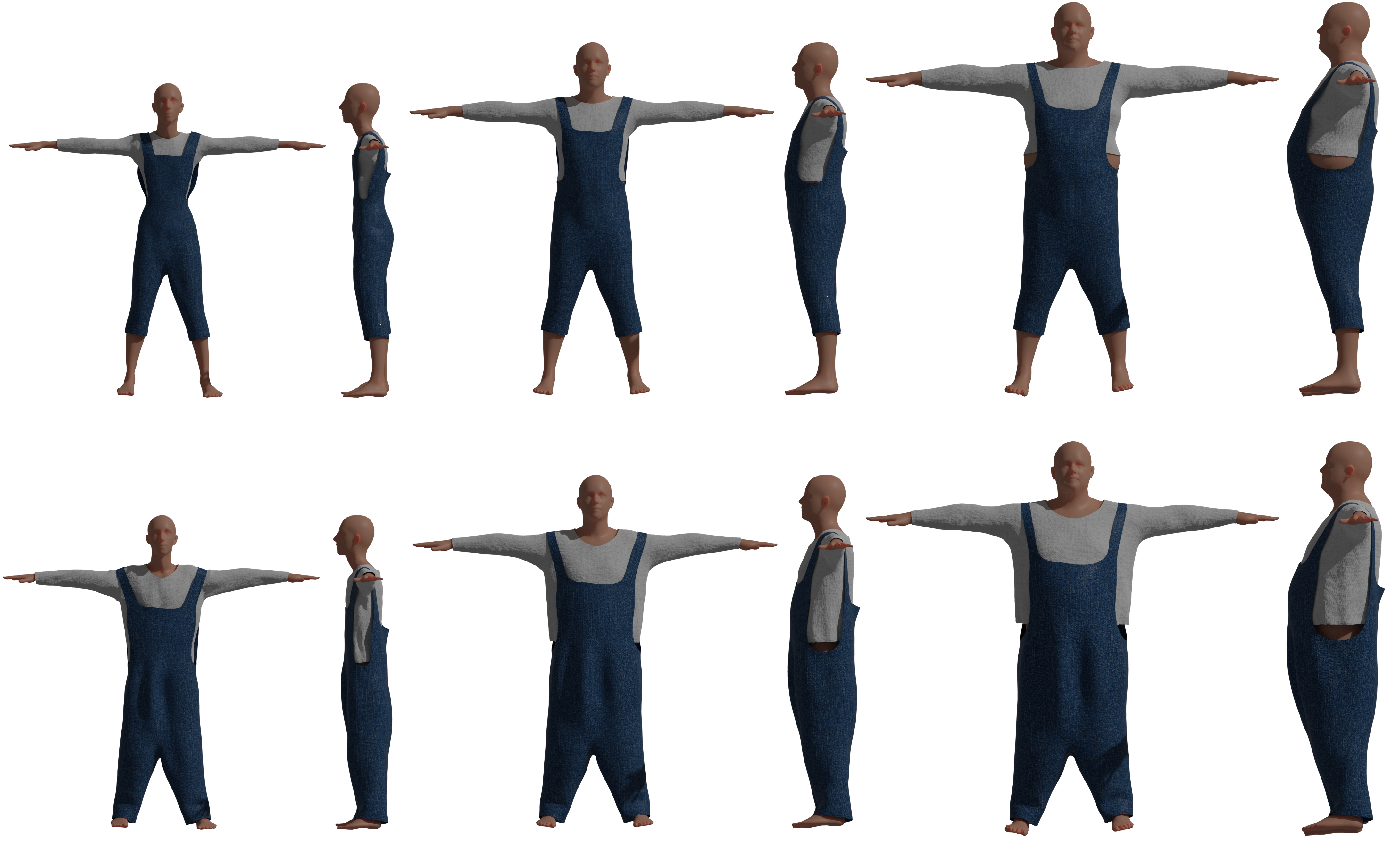}
    \caption{Results obtained using PBNS for outfit resizing. Body shape changes from left to right. Outfit tightness changes from upper to lower rows. All the previous properties discussed for PBNS are also present when it is used for outfit resizing.}
    \label{fig:resize}
\end{figure}

In the current literature, we observe an interest on automatic garment resizing \cite{vidaurre2020fully, tiwari2020sizer, patel2020tailornet}. Current approaches rely as well on data, which requires gathering, labelling, formatting and other issues already discussed. We observe that, again, the problem can be solved unsupervisedly using a standard 3D animation format. We propose to use PBNS to automatically obtain 3D \textit{animated} models --as blend shapes-- that morph into different body shapes and sizes. Note that given a body shape, the resulting outfit model should be able to be retargeted to this body, but also change in size. We refer to this concept as \textit{tightness}, while in previous works has been referred as \textit{style} \cite{patel2020tailornet}.

To this end, we replace PBNS input $\theta$ with the body shape $\beta$ concatenated with the garment tightness represented as $\gamma\in\mathbb{R}^2$. We also remove skinning. Then, we need to compute a prior approximation of the resized garment to have an estimate of $E_T$ in Eq.~\ref{eq:cloth}. To do so, we first transfer SMPL blend shapes to the outfit by proximity. To avoid carrying body geometry details, we exhaustively apply laplacian smoothing to these blend shapes ($100$ iterations). Finally, since garment shape variability is lower than body shape (except for tailored outfits), we keep only the first two blend shapes, corresponding to the first two shape parameters. Thus, we obtain blend shapes $\mathbf{D}^{2\times N\times3}$ to compute the estimation of $E_T$. We combine these blend shapes according to $\beta + \gamma$ (instead of just $\beta$), which shall yield $E_T$ for tighter or looser garments. Since PBNS is an unsupervised methodology and both, $\beta$ and $\gamma$ can be generated as uniform distributions within a reasonable range, it is possible to exhaustively train PBNS as a resizer for the whole input domain. Note that all PBNS properties discussed --multiple layers, fabrics, complements, etc.-- are also present when training for resizing. Fig.~\ref{fig:resize} contains the results obtained for a given outfit. We show three different body shapes (columns) and two different tightness (rows). As observed, PBNS allows resizing to the desired body shape and outfit tightness. The possibility to use PBNS as a methodology to obtain automatically resizable 3D outfits will also reduce 3D artist workload. Again, due to simplicity of the approach and its standard format --as blend shapes--, the engineering effort of integrating it into already existing 3D software is minimal, and its extreme efficiency (as standard PBNS, but without skinning) allows its application on real-scenarios.

\subsection{Custom Avatars}

\begin{figure*}
    \centering
    \includegraphics[width=\textwidth]{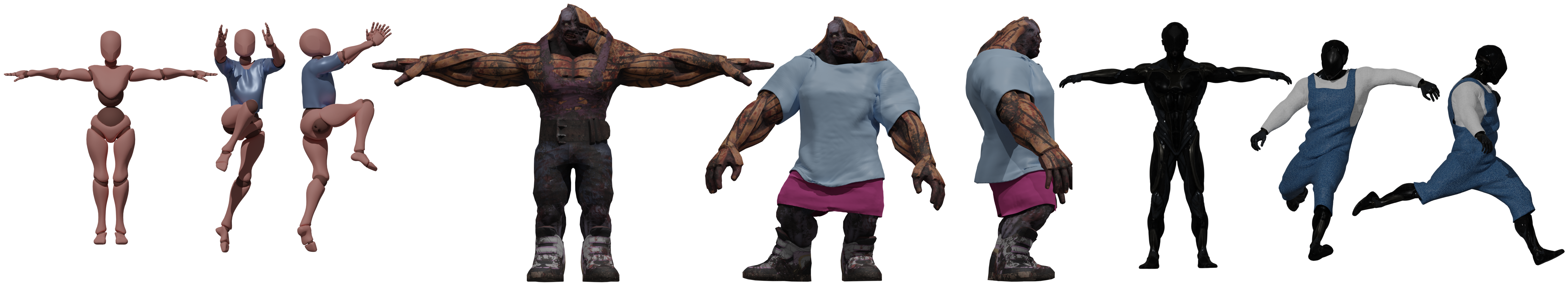}
    \caption{PBNS is not limited to work with SMPL. Since it needs no PBS data, the engineering effort for the adaptation to custom avatars is minimum. Here we show different garments and outfits neurally simulated on top of custom avatars. For each, we show the avatar body in rest pose and two different viewpoints of a posed sample. PBNS can be used to enhance rigged 3D characters by animating their clothes.}
    \label{fig:avatar}
\end{figure*}

PBNS formulation is not dependant of SMPL, and thus, it can be applied to animate outfits on top of any rigged 3D animated model. We gather different avatars and poses from Mixamo\footnote{https://www.mixamo.com/}, a free repository of 3D animated characters. We design or reuse outfits for the selected avatars. Then, we apply PBNS as described in this paper. Since no PBS data is required, we obtain animated outfits for the avatars in a matter of minutes. Fig.~\ref{fig:avatar} illustrates the results of this experiment. We show three different avatars. First, in rest pose, and then, two different viewpoints of the same pose. As can be seen, PBNS can be used to enhance rigged 3D characters by providing of realistic cloth behaviour to their outfits. This further proves the usefulness of the presented methodology for animated 3D character design.

\section{Performance}
\label{sec:performance}

\begin{table}[]
\centering
\begin{tabular}{ccc}
\hline
 & Single & Batch \\
\hline
CPU & $213$ FPS & $1235$ FPS \\
GPU & $455$ FPS & $14286$ FPS \\
\hline
\end{tabular}
\caption{PBNS performance. Tests done using an outfit with $12$k vertices and $23.7$k triangles, (the same as in Fig.~\ref{fig:vs_supervised}). We run the trained model in CPU and GPU for single and batched samples. As observed, the chosen architecture is extremely efficient. No related work (deep-based or PBS) comes close to this level of performance.}
\label{tab:performance}
\end{table}

We train our model on our subset of $2550$ training poses with a batch size of $16$ and Adam optimizer. We run our experiments on a GTX1080Ti. It takes around $1-2$ minutes per epoch, depending on the amount of collisions against the body. Using no GPU, it takes around $2-4$ minutes per epoch. Thus, training a model using the methodology presented in this paper does not require expensive hardware, enhancing accessibility for small film or videogame studios. Since there is no quantitative error, the stop criterion consists on qualitatively assessing validation predictions. It might take from $10$ to $50$ epochs to converge, depending on outfit complexity and body shape. During test, PBNS is extremely efficient. Tab.~\ref{tab:performance} shows the animation speed obtained with an outfit of $12$k vertices and $23.7$k triangles (same outfit as Fig.~\ref{fig:vs_supervised}). We run the model in both, CPU and GPU, for single and batched samples. As can be seen, our methodology can generate over $14$k samples per second. No previous work (deep-based or PBS) is near this level of performance. This is actually the expected behaviour, since PBNS yields skinned models with PSD. As aforementioned, these models are the standard for 3D animation and are designed to be extremely efficient. The only extra component is a small MLP that does not grow with vertex count. Since PBNS does not require post-processing, the reported numbers are the effective speed we would see in final applications. Additionally, due to the low size of the model, this solution is the only current methodology that can be applied in scenarios were computational resources must be available for other tasks, such as videogames and virtual reality, or in portable devices (see Tab.~\ref{tab:applicability}).

\section{Conclusions, Limitations and Future work}
\label{sec:conclusions}

We presented the first unsupervised deep learning based approach for outfit animation. More specifically, we described a methodology to neurally simulate outfits into blend shapes as Pose Space Deformations of Linear Blend Skinning models. Because of this, our solution is extremely efficient and can be easily integrated into any current 3D animation pipeline and run in almost any device (even low-computing or portable environments). We enabled unsupervised learning by formulating our problem as an implicit Physically Based Simulation. Furthermore, our proposed approach can be trained in a matter of minutes, even without a GPU. Therefore, the time from outfit design until model deployment is drastically reduced compared to previous approaches. PBNS can handle multiple layers of cloth, allowing neural simulation of complete outfits. Furthermore, we also show it can be easily adapted to any 3D avatar. This gives the methodology a broader applicability and higher scalability. CGI artists can design new animated draped characters more efficiently, and both, videogames and virtual try-ons, can easily introduce new 3D animated models for their outfit databases. Additionally, we presented the possibility of using this approach for automatic unsupervised outfit resizing.

In our approach, neither the input nor the potential energy formulation take into account the temporal dimension. This means that the learnt mapping from pose to mesh is unique. One can easily see how this is not true by imagining simulating the same pose sequence at different speeds. A given pose $\mathbf{\theta}$ on the sequence shall produce different meshes $\mathbf{V}_{\mathbf{\theta}}$. This is specially important for neural simulation of very loose garments (dresses, long skirts, etc.). Such garments present a very dynamic behaviour that static approaches cannot reproduce. We believe that including temporal behaviour in our neural simulator, while keeping its efficient formulation, is a promising research line as future work.

Finally, the idea of unsupervisedly learning to predict stable and physically consistent systems opens the possibility to generalize this methodology to handle other soft-tissue bodies. For example, hair, or human body self-collisions.

\textbf{Acknowledgements.} This work has been partially supported by the Spanish project PID2019-105093GB-I00 (MINECO/FEDER, UE) and CERCA Programme/Generalitat de Catalunya, by ICREA under the ICREA Academia programme, and by Amazon Research Awards.

\clearpage
{\small
\bibliographystyle{splncs04.bst}
\bibliography{egbib.bib}
}
\end{document}